\documentclass[nonacm,sigplan,screen]{acmart}

\newcommand{\Scout}{\textsc{Lever}}

\usepackage{amsmath}
\usepackage{algorithm}
\usepackage{algpseudocode}
\usepackage{makecell}
\usepackage{multirow}

\begin{document}

\title{\Scout{}: Speculative LLM Inference on Smartphones}

\author{Tuowei Wang}
\affiliation{%
  \institution{Tsinghua University}
  \city{Beijing}
  \country{China}
}

\author{Fengzu Li}
\affiliation{%
  \institution{Tsinghua University}
  \city{Beijing}
  \country{China}
}

\author{Yanfan Sun}
\affiliation{%
  \institution{Beihang University}
  \city{Beijing}
  \country{China}
}

\author{Wei Gao}
\affiliation{%
  \institution{University of Pittsburgh}
  \city{Pittsburgh}
  \country{USA}
}

\author{Ju Ren}
\authornote{Corresponding author (\href{mailto:renju@tsinghua.edu.cn}{renju@tsinghua.edu.cn}).}
\affiliation{%
  \institution{Tsinghua University}
  \city{Beijing}
  \country{China}
}

\begin{abstract}
Large language models (LLMs) are increasingly needed for interactive mobile applications, but high-quality models exceed the limited DRAM available on smartphones. Flash storage can hold larger models, yet flash-backed inference is slow because autoregressive decoding repeatedly invokes the target model and incurs costly I/O. We observe that speculative decoding is a natural fit for this setting: a small draft model can remain in DRAM, while a larger flash-resident target model verifies multiple candidate tokens per invocation. However, existing methods assume server-class accelerators and fail to account for prolonged I/O latency, limited computation parallelism, and irregular speculation execution.

We present \Scout{}, an end-to-end system for efficient flash-backed LLM inference on smartphones. \Scout{} jointly optimizes the three stages of speculative decoding under mobile constraints. For drafting, it builds token trees using an I/O- and compute-aware gain-cost objective. For verification, it prunes low-value branches through early-exit prediction to reduce target-model computation. For execution, it maps speculation efficiently across mobile CPU-NPU hardware to improve utilization. Comprehensive evaluations show that \Scout{} reduces inference latency by an average of $2.93\times$ over baseline flash-offloaded inference and $1.50\times$ over conventional speculative decoding, narrowing the latency gap between flash-backed and memory-resident LLM inference.
\end{abstract}

\maketitle
\pagestyle{plain} 

\section{Introduction}
As mobile applications become increasingly intelligent and interactive, there is a growing demand for running large language models (LLMs) directly on smartphones. Compared with cloud-based serving, on-device inference offers several important advantages. It improves privacy by keeping sensitive user data local~\cite{prism,kvshield}, avoiding the need to transmit personal context to remote servers. It also reduces serving cost~\cite{edge-shard,ce-collm}: instead of issuing repeated cloud requests for every interaction, inference can run directly on the device, making high-frequency workloads such as typing assistance and interactive editing more economical to support.

However, deploying high-quality LLMs on mobile devices remains fundamentally limited by memory. State-of-the-art models contain billions of parameters, while commodity smartphones provide only modest DRAM capacity. This DRAM is also shared with the operating system and other applications, leaving far less memory for LLM inference than the physical capacity suggests. Although compression techniques such as quantization~\cite{gptq,smoothquant,awq,bitnet} and pruning~\cite{llm-pruner,dejavu} can reduce model footprint, they may sacrifice generation quality and are often insufficient to fit larger, higher-quality models within mobile memory budgets.

A promising alternative is to exploit the heterogeneous storage hierarchy already available on smartphones~\cite{llm-flash,neuralink,powerinfer-2}. Flash storage offers substantially higher capacity than DRAM and can therefore hold model parameters that cannot reside in memory. By offloading model weights to flash, a device can support larger LLMs beyond DRAM capacity. However, this approach shifts the bottleneck from capacity to performance. During autoregressive decoding, the model is invoked repeatedly, once per generated token. If each invocation requires loading parameters from flash, I/O latency can dominate end-to-end inference time. Consequently, naively performing a flash-backed LLM inference yields latency far beyond what interactive mobile applications can tolerate.

This paper starts from the observation that \textit{speculative decoding is well matched to mobile DRAM-flash heterogeneous inference}. Specifically, speculative decoding separates LLM inference into two roles: a small speculative model (SSM) proposes draft tokens, while a larger target model verifies them. This separation aligns closely with the mobile memory hierarchy. The SSM can remain resident in DRAM and run with low latency, whereas the LLM can be stored in flash and invoked only intermittently. Instead of accessing the flash-resident LLM for every generated token, the system can use the SSM to generate multiple candidates and verify them with a single LLM invocation. When several candidates are accepted, the cost of loading and executing the LLM is amortized across several output tokens.

Despite this opportunity, existing speculative decoding methods~\cite{specinfer,eagle,medusa} are largely designed for server-class deployments, where the target model resides in high-bandwidth accelerator memory and verification can exploit abundant hardware parallelism. In these settings, verification is relatively inexpensive, shifting the optimization focus toward maximizing the acceptance rate of drafted tokens. Mobile inference breaks these assumptions. Because the target model is stored in flash, each verification step may incur expensive I/O and can take substantially longer than draft generation itself. Moreover, mobile systems-on-chip (SoCs) offer far less parallelism than server GPUs, so verification compute also becomes a major contributor to end-to-end latency. As a result, performance is no longer determined by speculation quality alone, but by how speculation interacts with mobile I/O and compute constraints. These constraints reshape the design space and expose three system-level challenges:

\noindent\textbf{Challenge \#1: Drafting.} The draft token tree must be redesigned for mobile flash-backed inference, where speculation interacts directly with both flash I/O and on-device compute constraints. If the tree is too small, the target model must be invoked frequently, causing repeated flash accesses and poor amortization of I/O latency. If the tree is too large, verification requires evaluating an excessive number of candidate tokens, introducing substantial compute overhead on resource-constrained mobile SoCs.

\noindent\textbf{Challenge \#2: Verification.} Fully verifying the tree can require substantial computation, especially when the tree contains many branches or candidate tokens. On mobile SoCs, where parallelism, memory bandwidth, and thermal headroom are limited, this verification step can easily become compute-bound. As a result, the latency saved by reducing flash-backed target invocations may be offset by the additional computation required for verification.

\noindent\textbf{Challenge \#3: Execution.} Mobile NPUs provide significant parallel compute capability, but they are optimized for regular tensor operations, predictable control flow, and hardware-aligned data layouts. Token-tree speculation, however, naturally introduces irregular branching, dynamic candidate structures, and fragmented computation. These properties can lead to inefficient kernel execution, poor data locality, and underutilization of NPU compute units.

\begin{figure}[t]
    \centering
    \includegraphics[width=1.0\linewidth]{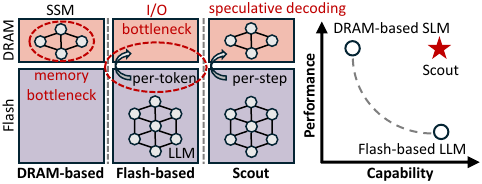}
    \caption{\Scout{} utilizes speculative decoding to combine the capacity of flash-based inference with the performance of DRAM-based inference on smartphones.}
    \label{fig:core-idea}
\end{figure}

To address these challenges, we present \Scout{}, an end-to-end LLM inference system based on speculative decoding with DRAM-flash heterogeneous storage on smartphones. As illustrated in Figure~\ref{fig:core-idea}, \Scout{} treats speculative decoding not merely as a decoding algorithm, but as a systems problem spanning drafting policy, verification cost, and mobile accelerator execution. \Scout{} consists of three components:

\noindent\textbf{Mobile-Optimized Drafting Construction.} \Scout{} constructs token trees by optimizing expected output tokens per speculative-cycle latency instead of acceptance rate alone. It estimates both the expected gain of candidate tokens and the mobile-specific cost of verifying them, then greedily expands high-value branches only when they are expected to improve end-to-end throughput. This allows \Scout{} to amortize expensive flash-backed target invocations without creating oversized trees that overload mobile compute.

\noindent\textbf{Predictor-Based Verification Pruning.} \Scout{} reduces target-model verification cost by pruning low-value branches before they traverse the full target model. It attaches a lightweight predictor to an intermediate target layer to estimate which draft branches are likely to survive final verification. To preserve correctness and avoid overly aggressive pruning, \Scout{} uses a conservative score-based algorithm to remove unlikely branches, while final token acceptance remains determined by the full target model.

\noindent\textbf{Hardware-Hybrid Execution Acceleration.} \Scout{} restructures speculative decoding to better match CPU-NPU execution on smartphones. During drafting, it batches selected branch expansions to improve NPU utilization without disrupting the gain-cost tree construction objective. During verification, it executes batched transformer computation on the NPU, but performs output projection selectively on the CPU only along the actual acceptance path, avoiding unnecessary logits computation for unvisited branches.

We evaluate \Scout{} under mobile-oriented DRAM-flash deployment settings, covering multiple models, tasks, and decoding configurations. The results show that \Scout{} achieves an average speedup of $2.93\times$ over flash-offloaded autoregressive inference. Compared with conventional speculative decoding, \Scout{} reduces latency by $1.50\times$ on average.

This paper makes the following contributions:
\begin{enumerate}
    \item We identify speculative decoding as a key opportunity for mobile DRAM-flash LLM inference, and reveal the system-level challenges required to make it practical.
    \item We present \Scout{}, a system that combines drafting, verification, and execution optimizations for efficient flash-backed LLM inference on smartphones.
    \item We extensively evaluate \Scout{}, showing consistent latency reductions over flash-backed inference and conventional speculative decoding methods.
\end{enumerate}

\section{Background}
\subsection{Memory Hierarchy on Smartphones}
\begin{figure}[t]
    \centering
    \includegraphics[width=1.0\linewidth]{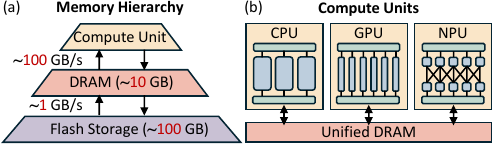}
    \caption{Smartphone hardware overview: (a) memory hierarchy and (b) heterogeneous compute units.}
    \label{fig:background-hardware}
\end{figure}

Modern smartphones expose a heterogeneous memory hierarchy consisting of DRAM and flash storage, as illustrated in Figure~\ref{fig:background-hardware}(a). DRAM provides low-latency access and relatively high bandwidth, making it the preferred location for storing model weights during inference. However, two factors limit the amount of DRAM that can be used by on-device LLMs. First, due to power and cost constraints, mobile devices typically have much smaller memory capacity than server-class systems, on the order of 10 GB. This capacity is insufficient for many modern LLMs, even when their weights are quantized. Second, the memory available to an LLM can be substantially smaller than the physical DRAM capacity, since DRAM is shared among the operating system, background services, and foreground applications. As a result, it is difficult to keep a high-quality multi-billion-parameter model fully resident in DRAM on smartphones.

Flash storage offers much larger capacity than DRAM and can therefore store model weights that exceed the available DRAM budget. This makes flash-backed inference a promising strategy for enabling larger LLMs on smartphones. However, flash storage provides much lower effective bandwidth than DRAM, typically 5 GB/s. Therefore, data transfers between flash and DRAM become a major performance bottleneck, particularly during autoregressive decoding, where weights are accessed repeatedly. As shown in Table~\ref{tab:preexp-flash-io}, if the target model must be loaded from flash for every generated token, I/O latency can dominate end-to-end inference time. Consequently, exploiting the large capacity of flash storage while minimizing inference latency is a fundamental challenge for an efficient mobile inference system.

\subsection{Compute Units on Smartphones}
As illustrated in Figure~\ref{fig:background-hardware}(b), smartphone SoCs integrate heterogeneous compute units, including CPUs, GPUs, and NPUs, which typically operate over a unified memory system. These compute units have distinct characteristics and are suited to different types of workloads. CPUs provide flexible control flow and are well-suited for system-level coordination and orchestration, but they offer limited throughput and energy efficiency for large-scale tensor operations. GPUs provide substantially higher parallel compute capability and can accelerate dense numerical workloads. On smartphones, however, GPUs are primarily designed for graphics rendering, so sustained GPU use for LLM inference may contend with foreground applications. In contrast, NPUs are specialized for neural network execution and provide better energy efficiency for dense tensor operations, making them an increasingly important backend for on-device LLM inference.

Nevertheless, mobile compute units remain far less capable and less flexible than server-class GPUs. This limitation is especially important for speculative decoding: although target-model verification can process multiple candidate tokens efficiently in parallel, the limited parallelism available on mobile devices diminishes this benefit. In addition, mobile NPUs typically execute compiled computation graphs, where operators, tensor shapes, and data layouts are optimized ahead of time. This execution model is suitable for regular tensor operations with static shapes, but is less effective for dynamic speculation patterns such as variable drafting structures and input-dependent verification paths.

\subsection{Speculative Decoding}
Autoregressive LLM inference is inherently sequential, as each token depends on previously generated tokens. Standard decoding is therefore severely memory-bound, since each step reads a large amount of model weights while performing limited computation for a single token. Speculative decoding~\cite{speculative-decoding,chen2023speculative,blockwise-parallel-decoding} accelerates this process by introducing two models with different roles: a lightweight \textit{draft model} proposes candidate tokens, and a larger \textit{target model} verifies them while defining the final output distribution. At each step, the draft model generates a candidate continuation, which the target model checks in one forward pass and accepts the longest prefix consistent with its own predictions. When multiple tokens are accepted, a single target-model invocation advances decoding by several positions, amortizing target-model cost across multiple output tokens.

\begin{figure}[t]
    \centering
    \includegraphics[width=1.0\linewidth]{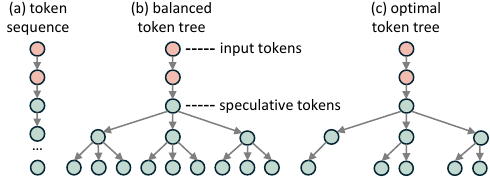}
    \caption{Speculative decoding topologies: (a) single sequence, (b) balanced tree, and (c) optimized tree (in \Scout{}).}
    \label{fig:background-speculative-decoding}
\end{figure}

Tree-based speculative decoding~\cite{specinfer,sequoia,eagle2,opttree} extends this idea by organizing draft candidates into a branching token tree rather than a single linear sequence. As depicted in Figure~\ref{fig:background-speculative-decoding}, instead of proposing only one continuation, the draft model explores multiple possible continuations from each position, and the target model verifies the entire tree in one forward pass. The accepted output then corresponds to a path in the tree that is consistent with the target model distribution. By covering multiple likely continuations at once, tree-based drafting can increase the expected number of accepted tokens. However, existing tree-based approaches primarily focus on algorithmic improvements and are typically studied in conventional server-side settings, whereas mobile inference introduces additional system-level constraints.

\section{Motivation and Challenges}
\subsection{Motivation}
Speculative decoding is a natural fit for LLM inference on smartphones with DRAM-flash heterogeneous storage. In this setting, a small speculative model can serve as the draft model and remain resident in DRAM, while the flash-backed LLM serves as the target model. Rather than loading the target model from flash and decoding every token individually, speculative decoding enables the DRAM-resident draft model to generate multiple tokens rapidly, invoking the target model only periodically for verification. When multiple candidates are accepted, the flash I/O and target-model execution costs are amortized over several output tokens, reducing average per-token latency. As a result, the I/O bottleneck between flash and DRAM on smartphones is alleviated.

Despite this opportunity, existing speculative decoding methods~\cite{speculative-decoding,specinfer,eagle,medusa,hydra,lookahead-decoding} are largely designed for server-class deployments. In such settings, both the draft and the target model reside in high-bandwidth memory, and abundant computation resources enable efficient parallel verification. Therefore, prior work has primarily focused on algorithm-level optimizations, such as improving draft quality and optimizing verification schedules. However, speculative LLM inference on smartphones differs in two important ways:

\begin{table}[b]
    \small
    \centering
    \caption{Per-token verification latency (ms) breakdown of flash-resident target models on a OnePlus 12.}
    \label{tab:preexp-flash-io}
    \begin{tabular}{lcccc}
        \toprule
        \textbf{Model} & \textbf{Compute} & \textbf{I/O} & \textbf{Total} & \textbf{I/O Percentage} \\ \midrule
        Llama-3.1-8B   & 93.2             & 990.6        & 1083.8         & 91.4\% \\
        Qwen3-4B       & 131.5            & 474.4        & 605.9          & 78.3\% \\
        Qwen3-8B       & 96.6             & 953.4        & 1050.0         & 90.8\% \\
        Qwen3-14B      & 141.2            & 1967.1       & 2108.3         & 93.3\% \\
        \bottomrule
    \end{tabular}
\end{table}

\noindent(1) \textbf{Flash-Bound Communication.} Given the limited DRAM capacity on smartphones, the target model is stored in flash, and each verification step requires loading its weights into DRAM. As shown in Table~\ref{tab:preexp-flash-io}, this data transfer can dominate verification latency due to the constrained effective bandwidth. By making verification more expensive, this additional communication alters the speculation cost trade-off and renders server-side drafting strategies suboptimal.

\begin{figure}[t]
    \centering
    \includegraphics[width=1.0\linewidth]{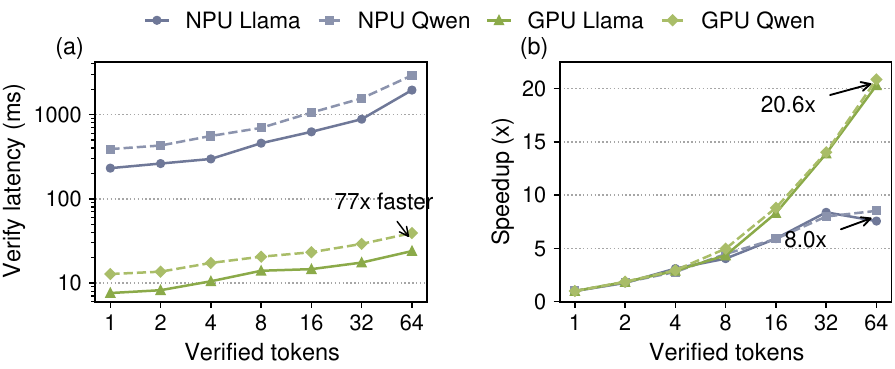}
    \caption{Verification parallelism on mobile and server hardware. We compare Llama-3.1-8B and Qwen3-8B on a Snapdragon 8 Gen 3 NPU and an NVIDIA RTX 3090 GPU. (a) Verification latency under different numbers of verified tokens. (b) Achieved speedup over single-token generation.}
    \label{fig:preexp-parallelism}
    \vspace{-0.5cm}
\end{figure}

\noindent(2) \textbf{Parallelism-Limited Computation.} Beyond memory limitations, mobile SoCs provide fewer computational resources than server GPUs. As shown in Figure~\ref{fig:preexp-parallelism}, verification latency on smartphones increases noticeably with the number of tokens being verified. Therefore, although speculative decoding can verify multiple drafted tokens in a single target-model pass, mobile accelerators are less effective at amortizing this cost through parallelism.

\subsection{Challenges}
These mobile-specific constraints reshape speculative decoding from a purely algorithmic optimization into a problem of algorithm-system co-design. Through a comprehensive analysis, we identify three critical technical challenges:

\begin{figure}[t]
    \centering
    \includegraphics[width=1.0\linewidth]{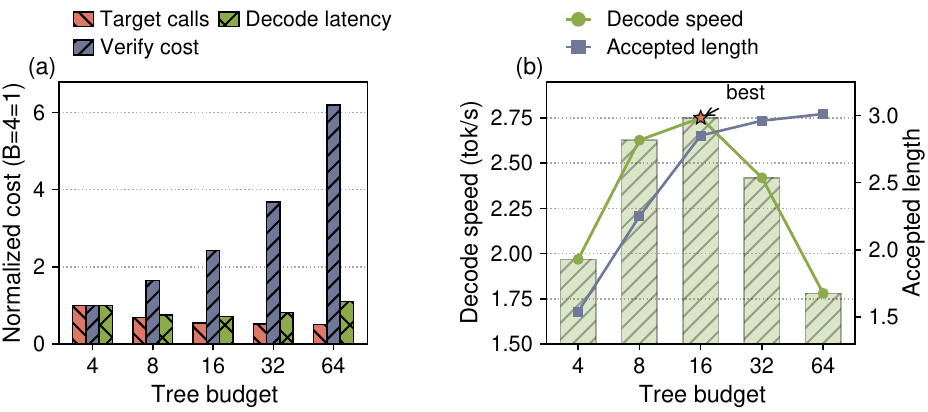}
    \caption{Token-tree budget dilemma on a OnePlus 12.
    (a) Normalized target-model calls, verification cost, and decoding cost under different tree budgets.
    (b) Decoding speed and accepted length under different tree budgets.}
    \label{fig:preexp-tree-budget}
\end{figure}

\noindent\textbf{Challenge \#1: Drafting.} Tree-based speculative decoding improves acceptance, but mobile systems require a redesigned draft token tree. A small token tree produces only a few candidates per verification step, causing frequent target-model invocations and poor I/O amortization. In contrast, an overly large token tree may accept more tokens, but it also increases the number of candidates that the target model must verify. As shown in Figure~\ref{fig:preexp-tree-budget}, mobile inference therefore faces a token-tree size dilemma: the drafting strategy must balance flash I/O accesses against verification compute cost, rather than optimizing acceptance rate alone.

\noindent\textbf{Challenge \#2: Verification.} Even with a carefully constructed token tree, each verification step requires computation across multiple branches and positions. On mobile SoCs, limited parallelism makes this verification step compute-bound. As a result, the latency saved from fewer flash-backed target invocations can be offset by the additional target-model computation required for full-tree verification. Moreover, pruning verification is difficult without compromising correctness, since every candidate that may affect the final accepted sequence must be faithfully checked.

\begin{table}[b]
    \small
    \centering
    \caption{Wasted verification computation across models and token tree budgets on a OnePlus 12.}
    \label{tab:preexp-lm-head}
    \begin{tabular}{lccc}
        \toprule
        \textbf{Target Model} & \textbf{Budget} & \textbf{Verify (ms)} & \textbf{Waste (ms / \%)} \\
        \midrule
        Llama-3.1-8B & 16 & 828  & 188 / 22.7 \\
        Llama-3.1-8B & 32 & 1255 & 256 / 20.4 \\
        Qwen3-8B     & 16 & 817  & 183 / 22.4 \\
        Qwen3-8B     & 32 & 1229 & 249 / 20.3 \\
        \bottomrule
    \end{tabular}
\end{table}

\noindent\textbf{Challenge \#3: Execution.} Efficient LLM inference on smartphones further requires fully exploiting available hardware resources. However, hardware characteristics and workload patterns are often mismatched, requiring careful adaptation. NPUs provide high throughput for regular tensor operations, but tree-based draft generation involves irregular branching that can underutilize them. In contrast, verification maps naturally to NPUs, yet their fixed execution patterns can introduce wasted computation on tokens that are ultimately not accepted. As shown in Table~\ref{tab:preexp-lm-head}, such wasted computation reaches 21.4\% on average and increases latency.

\section{System Design}
\subsection{Overview of \Scout{}}
\begin{figure}[t]
    \centering
    \includegraphics[width=1.0\linewidth]{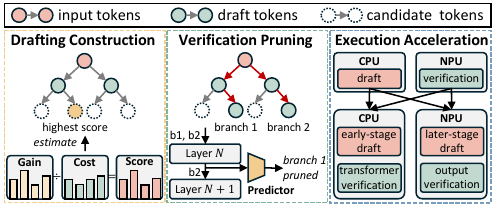}
    \caption{Overview of \Scout{}.}
    \label{fig:overview}
\end{figure}

We propose \Scout{}, an efficient LLM inference system that combines speculative decoding with DRAM-flash heterogeneous inference on smartphones. \Scout{} uses a lightweight speculative model resident in DRAM to accelerate inference for a large target model stored in flash. As illustrated in Figure~\ref{fig:overview}, \Scout{} introduces three key components that jointly optimize drafting, verification, and execution.

\noindent\textbf{Drafting Construction (\S~\ref{sec:drafting}).} During drafting, \Scout{} observes that the utility of each token depends on the current tree structure, along with both expected benefit and verification cost. To capture this trade-off, \Scout{} estimates the marginal gain and marginal cost of each token, and expands those with the highest gain-to-cost ratio. This enables \Scout{} to build compact token trees that prioritize high-value branches and maximize accepted tokens per unit latency.

\noindent\textbf{Verification Pruning (\S~\ref{sec:verification}).} For verification, \Scout{} leverages the fact that intermediate representations in the target model provide early signals of final output~\cite{deebert,fastbert,layerskip}. This allows low-value draft branches to be identified before traversing all target layers. \Scout{} inserts a lightweight predictor at an intermediate layer to score draft branches using hidden states. Branches with low predicted value are pruned, while accepted tokens are still decided by the full target model.

\noindent\textbf{Execution Acceleration (\S~\ref{sec:execution}).} At the execution layer, \Scout{} exploits complementary strengths of the CPU and NPU on smartphones. Rather than simply assigning drafting to the CPU and verification to the NPU, \Scout{} adopts a more hardware-aware strategy. It batches draft-model token generation to improve NPU utilization, and decouples output projection from NPU-based verification so that the CPU only computes logits along the actual acceptance path.

\subsection{Mobile-Optimized Drafting Construction}\label{sec:drafting}
\Scout{} formulates token-tree construction as an optimization problem that captures the trade-off between accepted-token throughput and mobile-specific costs. \Scout{} then estimates the key quantities in this objective through both analytical modeling and system profiling. Guided by these estimates, \Scout{} constructs the token tree using a greedy algorithm that can be efficiently executed at runtime. This design enables \Scout{} to generate token trees that are adapted to mobile constraints and optimized for low inference latency.

\noindent\textbf{Problem Formulation.} To maximize speculative decoding performance, \Scout{} adopts tree-based speculative decoding, where the draft is organized as a \textit{token tree} $T$. Each node represents a draft token, and each branch represents a candidate output path. The entire token tree can be verified efficiently by the target model. During verification, if a draft token matches the target model output, it is accepted. Otherwise, the draft token is rejected, and the target-model output is emitted as the fallback token of the current speculative cycle.

We use $\hat{G}(T)$ to denote the expected number of tokens produced in one speculative cycle, including both accepted draft tokens and the fallback token. We use $\hat{C}_{\mathrm{cycle}}(T)$ to denote the estimated latency of completing one speculative decoding cycle with tree $T$. The goal of construction is to make each cycle produce as many accepted tokens as possible under a limited latency budget. Based on this idea, \Scout{} selects the token tree with the largest gain per unit latency:
\begin{equation}\label{eq:overlap-objective}
T^{*}=\arg\max_{T}\frac{\hat{G}(T)}{\hat{C}_{\mathrm{cycle}}(T)}
\end{equation}

This objective directly corresponds to end-to-end decoding throughput. Compared with optimizing only for a larger number of accepted tokens, this formulation accounts for both accepted-token gain and additional system overhead.

\noindent\textbf{Key Quantity Estimation.} Before the token tree is constructed, its actual gain and cost are unknown. Therefore, \Scout{} estimates the key quantities in the objective using lightweight runtime statistics and system-aware profiling.

\noindent(1) \textit{Gain.} The gain of a token tree depends on how far target verification is expected to progress within the tree. For a non-root node $u$, let $\hat{b}(u)$ denote the estimated probability that the target verification path reaches $u$. Since $u$ contributes one accepted draft token exactly when it lies on the realized verification path, the expected number of accepted draft tokens is the sum of $\hat{b}(u)$ over all non-root nodes.

Therefore, the estimated gain of the token tree $T$ is:
\begin{equation}
\hat{G}(T)
=
1+
\sum_{u\in T\setminus\{\epsilon\}}
\hat{b}(u),
\label{eq:tree-gain}
\end{equation}
where $\epsilon$ denotes the root node. The summation term represents the expected number of accepted draft tokens, while the constant $1$ corresponds to the fallback token generated by the target model when verification stops.

\begin{figure}[t]
    \centering
    \includegraphics[width=1.0\linewidth]{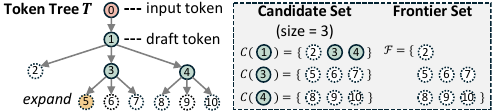}
    \caption{Key concepts in draft construction. Each expanded node generates a candidate set, and candidates whose parents are already in the tree form the frontier for greedy expansion.}
    \label{fig:design-drafting}
\end{figure}

As shown in Figure~\ref{fig:design-drafting}, for each node $u$, the \textit{candidate set} $C(u)$ consists of a fixed-size set of tokens with the highest draft-model probabilities conditioned on $u$. For a child $v=(u,a)$, where $a\in C(u)$ is a draft token proposed from parent $u$, \Scout{} estimates $\hat{b}(v)$ recursively as follows:
\begin{equation}
    \hat{b}(v)=\hat{b}(u)\cdot\operatorname{Calibrate}(p_d(a\mid u),C(u),r)
\end{equation}
where $p_d(a\mid u)$ is the draft probability of token $a$. The operator $\operatorname{Calibrate}(\cdot)$ maps this raw draft probability to the estimated probability that the target model also selects token $a$ after node $u$. It preserves the draft model's relative ranking within the candidate set, while using the reliability factor $r$ to adjust the overall confidence. The factor $r$ is updated from recent verification feedback: when target tokens often appear in the draft candidate sets, the calibrated value stays close to $p_d(a\mid u)$; otherwise, it is down-weighted to avoid overestimating unreliable branches.

\noindent(2) \textit{Cost.} \Scout{} estimates cost from two sources: the time required by the draft model to construct the tree and the time required by the target model to verify the tree:
\begin{equation}
\hat{C}_{\mathrm{cycle}}(T)
=
\hat{C}_{\mathrm{draft}}(T)
+
\hat{C}_{\mathrm{verify}}(|T|,L_T),
\label{eq:overlap-cost}
\end{equation}
where $|T|$ denotes the number of nodes entering the target verification batch, and $L_T$ denotes the number of leaf sequences induced by the tree.

The draft construction cost is relatively easy to estimate. Since the draft model is resident in memory and tree expansion is executed in small batches, \Scout{} directly maintains a lightweight moving average from runtime measurements to estimate the draft-side latency of each node or expansion batch. This avoids the need to manually build a complex analytical model for draft computation.

Target verification latency includes both verification computation and flash I/O. For a fixed target model and backend configuration, the flash I/O cost of one verification invocation is mainly determined by the target weights streamed from flash, and is therefore approximately fixed across different token-tree shapes within a cycle. The tree-dependent cost mainly comes from batched verification computation and runtime management. \Scout{} models this component using the verification shape $(|T|,L_T)$: $|T|$ determines the number of verification rows processed by the target model, while $L_T$ approximates the overhead of KV-cache copying, temporary sequence management, and sequence merging, because each leaf induces an independent temporary sequence.

To capture the combined verification cost, \Scout{} maintains an online latency profile indexed by $(|T|,L_T)$. This profile records the measured end-to-end verification latency after computation-I/O overlap and is queried during tree construction to estimate verification cost. For unseen shapes, \Scout{} estimates latency from nearby observed shapes with a conservative penalty. When the profile is empty, \Scout{} initializes it using offline measurements.

\begin{algorithm}[t]
\caption{Greedy Token Tree Construction}
\label{algo:token-tree}
\begin{algorithmic}[1]
\Require Context $x$, draft model $M_d$, and reliability factor $r$
\Ensure Token tree $T$
\State $T\gets\{\epsilon\}$, $\hat{b}(\epsilon)\gets 1$
\State $\hat{G}\gets 1$, $\hat{C}_{\mathrm{draft}}\gets 0$, $\mathcal{F}\gets\varnothing$
\State $\hat{C}_{\mathrm{draft}}\gets \hat{C}_{\mathrm{draft}}+\operatorname{AddFrontier}(\epsilon)$
\While{$\mathcal{F}\neq\varnothing$}
    \State $\hat{C}_{\mathrm{cycle}}\gets \hat{C}_{\mathrm{draft}}+\hat{C}_{\mathrm{verify}}(|T|,L_T)$
    \State $v^\star\gets \arg\max\limits_{v\in\mathcal{F}}\frac{\hat{b}(v)}{\Delta\hat{C}(v\mid T)}$
    \If{$\frac{\hat{b}(v^\star)}{\Delta\hat{C}(v^\star\mid T)}\le\frac{\hat{G}}{\hat{C}_{\mathrm{cycle}}}$}
        \State \textbf{break}
    \EndIf
    \State $T\gets T\cup\{v^\star\}$; $\mathcal{F}\gets\mathcal{F}\setminus\{v^\star\}$; $\hat{G}\gets\hat{G}+\hat{b}(v^\star)$
    \If{$v^\star$ is expandable}
        \State $\hat{C}_{\mathrm{draft}}\gets \hat{C}_{\mathrm{draft}}+\operatorname{AddFrontier}(v^\star)$
    \EndIf
\EndWhile
\State \Return $T$
\Procedure{AddFrontier}{$u$}
    \State $(C(u),t_u)\gets \operatorname{DraftCandidates}(M_d,x,u)$
    \ForAll{$a\in C(u)$ with draft probability $p_d(a\mid u)$}
        \State $v\gets(u,a)$
        \State $\hat{b}(v)\gets \hat{b}(u)\cdot\operatorname{Calibrate}(p_d(a\mid u),C(u),r)$
        \State $\mathcal{F}\gets\mathcal{F}\cup\{v\}$
    \EndFor
    \State \Return $t_u$
\EndProcedure
\end{algorithmic}
\end{algorithm}

\noindent\textbf{Greedy Tree Construction.} Based on the estimated quantities above, \Scout{} begins to construct the token tree. However, the exact optimization problem in Equation~\ref{eq:overlap-objective} is combinatorial, and solving it exactly would be too expensive to perform during decoding. \Scout{} adopts a lightweight greedy strategy, as outlined in Algorithm~\ref{algo:token-tree}.

As depicted in Figure~\ref{fig:design-drafting}, we use $\mathcal{F}$ to denote the \textit{frontier}, i.e., the set of generated candidate nodes whose parents are already in $T$ but which have not yet been inserted into the tree. This frontier differs from the candidate set $\mathcal{C}(u)$, which contains the draft tokens generated from a single node $u$.

Lines~1--3 initialize the tree with the root $\epsilon$, set the running gain to one fallback token, and create the initial frontier. The helper procedure \textsc{AddFrontier} generates the candidate set $\mathcal{C}(u)$ for a parent node $u$, estimates the reach probability of each generated child, and inserts these children into $\mathcal{F}$.

For a frontier node $v\in\mathcal{F}$, inserting $v$ into the current tree $T$ increases the expected gain by $\hat{b}(v)$, which is the estimated probability that the target verification path reaches $v$. It also changes the cycle latency, mainly due to the additional verification cost induced by the new tree shape and, if $v$ is expandable, the draft cost of generating its children. We denote this marginal latency increase by $\Delta\hat{C}(v\mid T)$.

At each iteration, Lines~5-6 compute the current cycle latency and select the frontier node with the largest marginal gain per unit latency as follows:
\begin{equation}
\label{eq:greedy-node-selection}
v^{\star}=\arg\max_{v\in\mathcal{F}}\frac{\hat{b}(v)}{\Delta \hat{C}(v\mid T)}
\end{equation}

After selecting $v^{\star}$, Lines~10-12 insert it into the token tree, invoke the draft model to generate the next-level candidate set $C(v^\star)$ if the node is expandable, and add these new candidate nodes to the frontier.

Finally, \Scout{} needs to decide when to stop expansion. The average gain rate of the current tree is
$\hat{G}(T) / \hat{C}_{\mathrm{cycle}}(T)$, which represents how many output tokens the tree is expected to produce per unit latency. The marginal gain rate of the best candidate in the frontier is
$\hat{b}(v) / \Delta\hat{C}(v\mid T)$. If even the best remaining candidate cannot exceed the average gain rate of the current tree, adding it would only reduce the overall gain per unit latency. Therefore, Lines~7-8 stop expansion when even the best remaining node cannot improve the current average gain rate:
\begin{equation}
\max_{v\in\mathcal{F}}
\frac{\hat{b}(v)}{\Delta\hat{C}(v\mid T)}
\le
\frac{\hat{G}(T)}{\hat{C}_{\mathrm{cycle}}(T)}.
\label{eq:overlap-stop}
\end{equation}
This condition serves as a stopping criterion to avoid adding nodes with low marginal gain. It allows \Scout{} to avoid relying on a fixed tree budget to determine the tree size.

\subsection{Predictor-Based Verification Pruning}\label{sec:verification}
Due to the limited compute capability of mobile devices, fully verifying a branched token tree can easily become a new latency bottleneck. To reduce this overhead, \Scout{} introduces predictor-based pruning during verification. By accurately identifying and removing low-value branches at an intermediate layer of the target model, \Scout{} reduces verification computation while preserving output quality.

\noindent\textbf{Early-Exit Predictor.} \Scout{} introduces a lightweight neural-network predictor at a middle layer of the target model. Given the hidden states of tree nodes, the predictor estimates whether each branch is worth continuing through the remaining layers. Branches predicted to have low value are terminated early and removed from subsequent verification.

\noindent(1) \textit{Model Structure.} The predictor is implemented as a linear projection $W \in \mathbb{R}^{|\mathcal{V}|\times d}$, where $|\mathcal{V}|$ is the vocabulary size and $d$ is the hidden-state dimension. Each row of $W$ corresponds to the scoring vector of one token. The predictor takes the intermediate hidden state from the target model as input and outputs a score for each candidate token.

When verification reaches layer $L$, each node sent to the target model obtains an intermediate hidden state. Let $h_u^L$ denote the hidden state of node $u$ at layer $L$. For a node $u$ and its candidate token $a$, the predictor reads the corresponding row $W[a]$ and computes the score as follows:
\begin{equation}
s(u,a)=W[a]^\top h_u^L.
\label{eq:early-exit-logit}
\end{equation}
This score estimates whether token $a$ is worth preserving based on the hidden state of its parent node.

\noindent(2) \textit{Training Process.} During offline training, the target model is frozen and only the predictor is updated. For each training position $t$, \Scout{} records the intermediate hidden state $h_t^L$ and the final logits $z_t$ produced by the target model. To encourage the predictor to approximate the final output distribution of the target model, the training objective consists of two components: a full-vocabulary knowledge-distillation loss $\mathcal{L}_{\mathrm{KD}}$ and a candidate-set distillation loss $\mathcal{L}_{\mathrm{Cand}}$. The final training objective is defined as:
\begin{equation}
\label{eq:early-exit-loss}
\mathcal{L}=\mathcal{L}_{\mathrm{KD}}+\lambda_{\mathrm{Cand}}\mathcal{L}_{\mathrm{Cand}},
\end{equation}
where $\lambda_{\mathrm{Cand}}$ controls the strength of candidate-set alignment.

The first term is a knowledge-distillation loss. The final-layer softmax distribution of the target model serves as the teacher, while the early-exit distribution produced by the predictor serves as the student. This encourages the intermediate predictor to approximate the final distribution of the target model without modifying the target model itself:
\begin{equation}
\mathcal{L}_{\mathrm{KD}}
=
\tau_{\mathrm{KD}}^2
\sum_t
\mathrm{KL}
\left(
\mathrm{softmax}\left(\frac{z_t}{\tau_{\mathrm{KD}}}\right)
\;\middle\|\;
\mathrm{softmax}\left(\frac{W h_t^L}{\tau_{\mathrm{KD}}}\right)
\right),
\label{eq:early-exit-kd-loss}
\end{equation}
where $\tau_{\mathrm{KD}}$ is the distillation temperature.

Furthermore, \Scout{} adds a candidate-set distillation loss because inference-time pruning is not a full-vocabulary prediction problem. The predictor only needs to solve a local ranking problem: among the candidate tokens proposed by the draft model, it should identify which ones are more likely to match the final decision of the target model. Specifically, at each training position $t$, the draft model produces a candidate set of node $u$ $C_{t}(u)$. Let $q_T(\cdot\mid t)=\operatorname{softmax}_{C_{t}(u)}(z_t/\tau_{\mathrm{Cand}})$ and $q_E(\cdot\mid t)=\operatorname{softmax}_{C_{t}(u)}(W h_t^L/\tau_{\mathrm{Cand}})$ denote the target and early-exit distributions normalized only over $C_{t}(u)$. The resulting candidate-set distillation loss is:
\begin{equation}
\mathcal{L}_{\mathrm{Cand}}
=
\tau_{\mathrm{Cand}}^2
\sum_t
\mathrm{KL}
\left(
q_T(\cdot \mid t)
\;\middle\|\;
q_E(\cdot \mid t)
\right).
\label{eq:early-exit-cand-loss}
\end{equation}

\begin{figure}[t]
    \centering
    \includegraphics[width=1.0\linewidth]{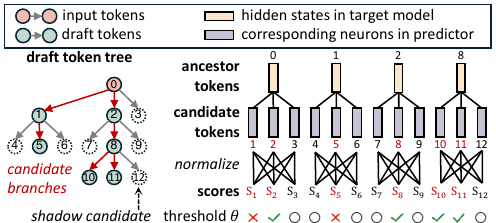}
    \caption{\Scout{} uses intermediate hidden states from the target model and a lightweight predictor to score candidate branches. Scores are normalized within each candidate set, including shadow candidates, and low-value branches are pruned before completing full target-model verification.}
    \label{fig:design-verification}
\end{figure}

\noindent\textbf{Score-Based Algorithm.} Based on the trained predictor, \Scout{} scores the nodes and their candidate tokens in the token tree to decide which branches can be pruned, as shown in Figure~\ref{fig:design-verification}. A threshold $\theta$ controls the pruning trade-off. If a token receives a score higher than $\theta$, it is likely to match the final choice of the target model, making the corresponding edge valuable to preserve. Conversely, if a token is unlikely to be accepted by the target model, allowing it to pass through the remaining layers only wastes computation.

\noindent(1) \textit{Candidate Normalization.} Instead of directly using the raw predictor score, \Scout{} first normalizes scores within each candidate set. This design is important when a parent node has only one inserted child: without additional comparison targets, that child would always receive the highest score, making the pruning decision meaningless. To address this issue, \Scout{} scores not only the actual children in the token tree but also the draft-model candidate set $C(u)$. Candidate tokens that are not inserted into the token tree are called \textit{shadow tokens}; they serve only as comparison references.

\Scout{} normalizes the scores within the candidate set of each node as follows:
\begin{equation}\label{eq:early-exit-score}
s_{norm}(u,a) =
\frac{\exp(s(u,a)/\tau_E)}
{\sum_{b\in C(u)}\exp(s(u,b)/\tau_E)},
\end{equation}
where $\tau_E$ is the temperature parameter. This score is not a target probability over the full vocabulary. Instead, it is a local relative score: among the candidates proposed by the draft model for parent node $u$, it measures how valuable it is to continue verifying token $a$ according to the predictor. Shadow tokens appear only in the denominator. They are not inserted into the token tree, do not enter subsequent target verification, and can never be directly accepted as output tokens. Their only role is to make the comparison meaningful for actual tree edges, including edges whose parent has only one inserted child.

\noindent(2) \textit{Conservative Pruning.} A basic threshold rule marks an edge $(u,a)$ as removable when$s_{norm}(u,a)<\theta$, but the final keep set must satisfy several safeguards. First, if a node is kept, all of its ancestors must also be kept. Second, \Scout{} preserves a backbone path. Starting from the root, this path repeatedly selects the most reliable child at each depth until reaching either a leaf node or the maximum depth. The backbone path preserves a main progress path and prevents the predictor from eliminating the depth gain of the entire tree due to a local misprediction. Near the root, where next-token uncertainty is often higher, \Scout{} may keep multiple high-scoring children to preserve necessary branch coverage. Finally, if pruning leaves too few nodes or too few leaves, \Scout{} rejects the pruning decision as overly aggressive.

After the keep decisions are made, \Scout{} compacts the remaining nodes into a new tree $T'$. The target model then continues from layer $L$ to the final layer only for the nodes preserved in $T'$. Therefore, for any accepted path in $T'$, the final token decisions remain aligned with the target model, preserving output quality while reducing computation.

\subsection{Hardware-Hybrid Execution Acceleration}\label{sec:execution}
\begin{figure}[t]
    \centering
    \includegraphics[width=1.0\linewidth]{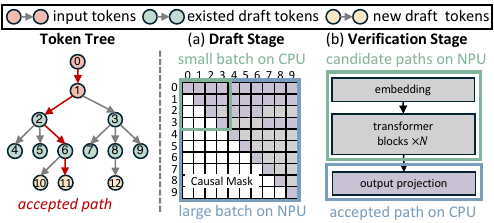}
    \caption{(a) During drafting, \Scout{} schedules small dynamic expansions on the CPU and batches larger regular expansions on the NPU. (b) During verification, transformer computation is executed as batched NPU work, while output projection is performed on demand on the CPU only along the accepted path to avoid redundant logits computation.}
    \label{fig:design-execution}
    \vspace{-0.3cm}
\end{figure}

As illustrated in Figure~\ref{fig:design-execution}, \Scout{} adopts CPU-NPU cooperative execution and optimizes the two stages of speculative decoding separately. In the draft stage, \Scout{} improves NPU utilization through batch-aware scheduling without violating the construction objective of the token tree. In the verification stage, \Scout{} places the transformer layers that are suitable for batched execution on the NPU, while moving the output projection, which is only needed along the actual accepted path, to the CPU and executing it on demand.

\noindent\textbf{Draft: Batch-Aware Scheduling.} The draft stage is typically executed on the CPU, because NPUs are optimized for tensor computation with regular shapes and sufficient arithmetic intensity. On the one hand, the draft stage involves the dynamic construction of a token tree: whether each branch should be extended is determined online based on the current tokens. On the other hand, conventional token trees are usually small and therefore do not provide enough computation to fully utilize the NPU. In such cases, additional kernel launch and data transfer overheads can dominate latency, offsetting the benefit of NPU execution.

However, due to I/O bottlenecks and limited compute capability on smartphones, verification is substantially more time-consuming than drafting. Consequently, the token tree is expanded to a larger size and gradually accumulates multiple high-value nodes. Since the parents of these nodes are already included in the tree, they can be expanded together to generate next-level candidates. For such nodes, the draft model performs the same forward computation, differing only in the input node. With an appropriate causal mask, \Scout{} can organize these computations into a batch and execute them efficiently on the NPU.

\Scout{} designs a batch-aware scheduling strategy. In the early phase of token tree construction, when the tree width is small and the computation shape changes frequently, \Scout{} executes nodes serially on the CPU. As the tree grows and more high-value candidates become available, \Scout{} groups these candidates into batches and executes them on the NPU. Through this progressive scheduling strategy, \Scout{} exploits the batching capability of the NPU while preserving the gain-cost optimization objective of token tree construction.

\noindent\textbf{Verification: Layer-Wise Separation.} In contrast to the draft stage, the verification stage is naturally suitable for NPU execution. After pruning, the remaining draft tokens can be organized into a compact batch for computation. However, because the vocabulary size is large, the output projection often accounts for a substantial portion of verification latency and becomes a bottleneck that is difficult to hide through layer-wise compute/I/O overlap. Moreover, NPUs on smartphones typically rely on a predefined compute graph with a fixed batch shape. When the number of candidate paths to be verified is large, all paths must participate in the full output projection, even though only one path will eventually be accepted and the logits of the other paths will never be used. This introduces significant redundant computation.

To address this issue, \Scout{} separates the verification stage into two parts. The first part is transformer computation. For the nodes preserved after pruning, \Scout{} still organizes them into a verification batch and executes all transformer layers on the NPU. The second part is logits generation. After the transformer layers finish, \Scout{} keeps only the final hidden state of each node. Since subsequent logits computation depends on the actual path taken by target sampling, \Scout{} removes this part from the fixed NPU graph and executes it on demand on the CPU.

Specifically, after executing the target transformer layers and obtaining the final hidden state of each token, \Scout{} first computes logits only for the root token and performs target sampling. If the sampled token matches a candidate token, \Scout{} then applies the output projection to the hidden state of that candidate. This process proceeds step by step along the accepted path until the sampled token no longer matches any candidate or the end of the path is reached. By eliminating logits computation for a large number of unaccepted paths, \Scout{} reduces end-to-end verification latency even though this part is executed on the CPU.

\section{Evaluation}
\subsection{Experimental Setup}
\noindent\textbf{Hardware.} We evaluate \Scout{} on three commercial smartphones, as summarized in Table~\ref{tab:eval-hardware}. Unless otherwise stated, all ablation studies and sensitivity analyses are conducted on the OnePlus 12. The overall end-to-end evaluation is conducted on all three devices.

\begin{table}[t]
    \small
    \centering
    \caption{Hardware Configuration.}
    \label{tab:eval-hardware}
    \begin{tabular}{lll}
        \toprule
        \textbf{Device} & \textbf{SoC}       & \textbf{Flash Storage} \\ \midrule
        OnePlus 12      & Snapdragon 8 Gen 3 & UFS 4.0                \\
        Honor 500 Pro   & Snapdragon 8 Elite & UFS 3.1                \\
        Honor Magic V5  & Snapdragon 8 Elite & UFS 4.1                \\
        \bottomrule
    \end{tabular}
\end{table}

\begin{table}[t]
    \centering
    \caption{Model Configuration.}
    \label{tab:eval-models}
    \small
    \begin{tabular}{llll}
        \toprule
        \textbf{Model} & \textbf{Prompt Mode} & \textbf{Model} & \textbf{Prompt Mode} \\ \midrule
        Llama-3.1-8B   & Instruct             & Qwen3-8B       & Thinking             \\
        Qwen3-4B       & Non-Thinking         & Qwen3-14B      & Thinking             \\
        \bottomrule
    \end{tabular}
\end{table}

\begin{figure*}[t]
    \centering
    \includegraphics[width=1.0\linewidth]{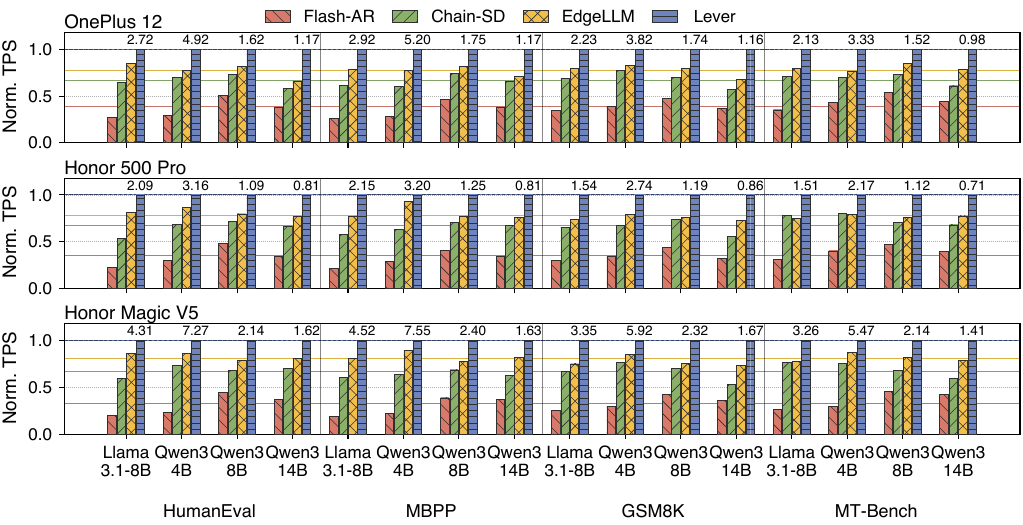}
    \caption{End-to-end decode throughput across all 48 device-model-dataset configurations, normalized to \Scout{}. Labels above the \Scout{} bars indicate absolute throughput in tokens/s.}
    \label{fig:eval-overall}
\end{figure*}

\noindent\textbf{Models.} Table~\ref{tab:eval-models} lists the target models used in the evaluation. For speculative decoding, we adopt EAGLE3~\cite{eagle3}, one of the most widely used draft-model training methods, to construct the draft model paired with each target model. We use the same target-draft pairs across \Scout{} and all speculative baselines for fairness, and quantize both target and draft models to Q4\_0. Unless otherwise specified, target-model weights are stored in flash and streamed layer by layer during decoding. For Qwen3 models, Thinking and Non-Thinking denote the prompt settings that enable or disable the model's reasoning mode, respectively; for Llama-3.1, Instruct denotes the instruction-tuned chat template.

\noindent\textbf{Datasets.}
We evaluate on four benchmarks: HumanEval~\cite{humaneval}, MBPP~\cite{mbpp}, GSM8K~\cite{gsm8k}, and MT-Bench~\cite{mtbench}. These datasets cover code generation, mathematical reasoning, and open-ended dialogue workloads. We use each benchmark to provide representative decoding workloads rather than to evaluate task accuracy. For each dataset, we randomly select 50 samples and measure generation speed. Unless otherwise specified, the maximum output length is 128 tokens. Although \Scout{} supports stochastic sampling, our main evaluation uses greedy decoding with temperature 0 to eliminate generation randomness and ensure reproducible comparisons. Our primary metric is decode throughput in tokens per second. We exclude prefill time and measure only the decode phase. For aggregate results, we report the geometric mean across models, datasets, and devices.

\noindent\textbf{Baselines.}
We compare \Scout{} with three different baselines:
\begin{itemize}
    \item \textbf{Flash-AR} is standard autoregressive decoding with a flash-backed target model. At each decoding step, it streams the target-model weights from flash layer by layer, runs the target model once to generate a single token, and then proceeds to the next step.
    \item \textbf{Chain-SD} uses standard sequence-based speculative decoding~\cite{speculative-decoding} with the same EAGLE3 draft model as \Scout{}. In each cycle, it drafts a fixed 8-token chain and verifies the chain with one flash-backed target-model invocation, accepting the longest valid prefix.
    \item \textbf{EdgeLLM}~\cite{edgellm} applies a similar speculative decoding approach to accelerate flash-backed LLM inference on mobile devices. Compared with \Scout{}, EdgeLLM focuses primarily on algorithm-level optimizations, including branch navigation and verification pacing.
\end{itemize}

In addition to these end-to-end baselines, we include SpecInfer~\cite{specinfer} and OPT-Tree~\cite{opttree} only in the token-tree construction ablation. SpecInfer represents the tree-based speculative inference, while OPT-Tree optimizes the token-tree structure to improve expected acceptance.

For fairness, all methods use the same target model, quantization format, prompt template, decoding configuration, and flash-backed weight-streaming mechanism. Speculative methods additionally use the same EAGLE3 draft model.

\subsection{End-to-End Performance}

Figure~\ref{fig:eval-overall} reports end-to-end decode throughput across all 48 configurations. \Scout{} achieves a geometric-mean speedup of 2.93$\times$ over Flash-AR, 1.50$\times$ over Chain-SD, and 1.27$\times$ over EdgeLLM. The gain over Flash-AR comes from amortizing flash-backed target-model invocations, while the gains over Chain-SD and EdgeLLM show the importance of adapting speculation to mobile flash I/O, limited verification parallelism, and CPU-NPU execution costs. \Scout{} consistently outperforms all the baselines. Its gains are larger on code-generation and reasoning workloads, where draft continuations are more reliable, and smaller on MT-Bench, where open-ended dialogue leads to weaker draft-target agreement. These results show that \Scout{} generalizes across hardware, models, and workloads.

\subsection{Ablation Study}

\begin{figure}[t]
    \centering
    \includegraphics[width=1.0\linewidth]{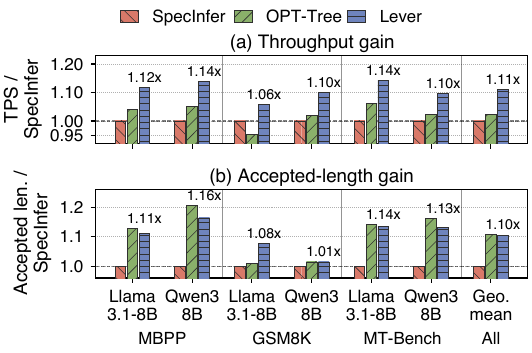}
    \caption{Decode throughput and average accepted length across draft policies. Values are normalized to SpecInfer.}
    \label{fig:eval-drafting}
\end{figure}

\noindent\textbf{Mobile-Optimized Drafting Construction.} Figure~\ref{fig:eval-drafting} compares different draft policies. Only the construction policy has changed. Compared with SpecInfer, \Scout{} achieves both higher decode throughput and longer average accepted length, showing that its gain estimate selects more useful speculative branches. Compared with OPT-Tree, \Scout{} has slightly lower accepted length but higher throughput. This shows that maximizing accepted length alone can introduce extra verification overhead on mobile hardware. By optimizing accepted tokens per unit latency, \Scout{} builds a more cost-effective token tree and improves decode speed.

\begin{figure}[t]
    \centering
    \includegraphics[width=1.0\linewidth]{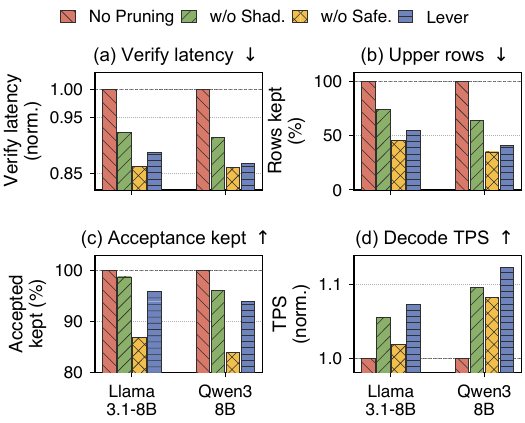}
    \caption{Effect of pruning policies on verification latency, upper-layer rows kept, accepted-token retention, and decode throughput, normalized to no pruning where applicable.}
    \label{fig:eval-pruning}
\end{figure}

\noindent\textbf{Predictor-Based Verification Pruning.} Figure~\ref{fig:eval-pruning} compares pruning policies using the same token tree and execution backend. A verification row corresponds to a token-tree node in the target verification batch; rows kept denotes the fraction of nodes that continue after the early-exit layer, and accepted-token retention denotes the accepted length relative to full verification on the same tree. \Scout{} keeps about 48\% of upper-layer rows, reduces verification latency by 12.2\%, and still retains about 95\% of accepted tokens.

The results show that both components are necessary. Without shadow candidates, scores are less reliable when a parent has few children, limiting pruning opportunities. Without safeguards, pruning removes more rows but also discards useful paths. Overall, \Scout{} reduces upper-layer verification work while preserving most accepted progress.

\begin{figure}[t]
    \centering
    \includegraphics[width=1.0\linewidth]{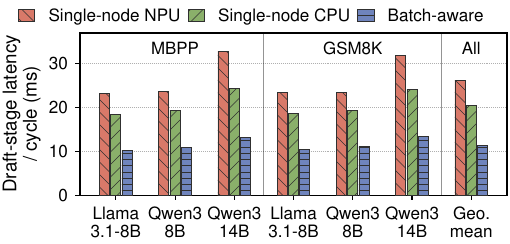}
    \caption{Draft-stage latency under Single-node NPU, Single-node CPU, and batch-aware NPU scheduling.}
    \label{fig:eval-draft-scheduling}
\end{figure}

\begin{figure}[t]
    \centering
    \includegraphics[width=1.0\linewidth]{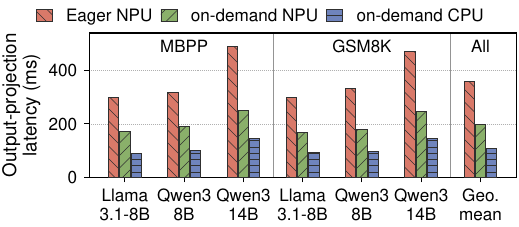}
    \caption{Output-projection latency under eager NPU, on-demand NPU, and on-demand CPU projection.}
    \label{fig:eval-lazy-output}
\end{figure}

\noindent\textbf{Hardware-Hybrid Execution Acceleration.} Figure~\ref{fig:eval-draft-scheduling} compares different schedules for draft expansion. Single-node NPU execution suffers from launch and synchronization overhead, while Single-node CPU execution cannot exploit tensor parallelism. Batch-aware scheduling groups high-value frontier expansions into regular NPU batches and reduces draft-stage latency without changing the token-tree construction policy. Figure~\ref{fig:eval-lazy-output} compares where and when output projection is executed. Eager NPU projection computes logits for all verified nodes, including branches that are never visited. On-demand projection computes logits only when target sampling reaches a node on the accepted path. Compared with on-demand NPU projection, on-demand CPU projection avoids repeated small NPU invocations and CPU-NPU synchronization, reducing output-projection latency while producing the same accepted tokens.

\subsection{Sensitivity Analysis}

\begin{table}[t]
    \small
    \centering
    \caption{Decode throughput on Qwen3-8B with different draft models. Entries report decode speed (tokens/s).}
    \label{tab:eval-draft-quality}
    \setlength{\tabcolsep}{3pt}
    \begin{tabular}{llllll}
        \toprule
        \textbf{Draft} & \textbf{Method} &
        \textbf{MBPP} & \textbf{GSM8K} & \textbf{MT-Bench} & \textbf{Geo.} \\
        \midrule
        \multirow{3}{*}{EAGLE3}
        & Chain-SD & 1.296 & 1.222 & 1.108 & 1.206 \\
        & EdgeLLM  & 1.434 & 1.382 & 1.283 & 1.365 \\
        & \Scout{} & \textbf{1.753} & \textbf{1.741} & \textbf{1.516} & \textbf{1.666} \\
        \midrule
        \multirow{3}{*}{\makecell{Qwen3\\(0.6B)}}
        & Chain-SD & 0.860 & 0.840 & 0.820 & 0.840 \\
        & EdgeLLM  & 1.010 & 0.940 & 0.890 & 0.945 \\
        & \Scout{} & \textbf{1.250} & \textbf{1.220} & \textbf{1.110} & \textbf{1.192} \\
        \bottomrule
    \end{tabular}
\end{table}

\begin{table}[t]
    \small
    \centering
    \caption{Decode throughput as the fraction of target-model layers resident in DRAM varies. Entries report tokens/s. L8 and Q8 denote Llama-3.1-8B and Qwen3-8B, respectively.}
    \label{tab:eval-dram-budget}
    \begin{tabular}{lllllll}
        \toprule
        \textbf{Workload} &
        \textbf{0\%} &
        \textbf{20\%} &
        \textbf{40\%} &
        \textbf{60\%} &
        \textbf{80\%} &
        \textbf{100\%} \\
        \midrule
        Q8 / MBPP   & 1.753 & 2.010 & 2.473 & 2.812 & 3.120 & 3.401 \\
        L8 / MBPP   & 2.918 & 3.704 & 4.214 & 4.619 & 4.922 & 5.494 \\
        Q8 / GSM8K  & 1.741 & 1.996 & 2.453 & 2.791 & 3.099 & 3.377 \\
        L8 / GSM8K  & 2.225 & 2.821 & 3.203 & 3.508 & 3.736 & 4.194 \\
        \bottomrule
    \end{tabular}
\end{table}

\begin{table}[t]
    \small
    \centering
    \caption{Decode throughput under different maximum output lengths. Entries report tokens/s. L8 and Q8 denote Llama-3.1-8B and Qwen3-8B, respectively.}
    \label{tab:eval-output-length}
    \begin{tabular}{llllll}
        \toprule
        \textbf{Workload} &
        \textbf{32} &
        \textbf{64} &
        \textbf{128} &
        \textbf{256} &
        \textbf{512} \\
        \midrule
        Q8 / MBPP   & 1.768 & 1.772 & 1.753 & 1.741 & 1.712 \\
        L8 / MBPP   & 3.082 & 3.110 & 2.918 & 2.890 & 2.854 \\
        Q8 / GSM8K  & 1.756 & 1.760 & 1.741 & 1.729 & 1.681 \\
        L8 / GSM8K  & 2.349 & 2.371 & 2.225 & 2.204 & 2.173 \\
        \bottomrule
    \end{tabular}
\end{table}

\noindent\textbf{Draft-model quality.} Table~\ref{tab:eval-draft-quality} evaluates \Scout{} with different draft models for the same Qwen3-8B target. We compare the EAGLE3 draft model with a smaller Qwen3-0.6B draft model to test whether the design depends on a highly specialized draft. Although the weaker draft reduces absolute throughput for all speculative methods, \Scout{} consistently outperforms Chain-SD and EdgeLLM in both settings. This suggests that \Scout{} generalizes across draft-model choices, rather than relying on one particular draft model.

\noindent\textbf{DRAM budget.} Table~\ref{tab:eval-dram-budget} varies the fraction of target-model layers pinned in DRAM. A budget of $x\%$ means that the first $x\%$ of target layers remain resident, while the rest are streamed from flash during each verification step. Throughput improves as the budget increases because each target invocation loads fewer weights from flash. More importantly, \Scout{} remains effective across all budgets, including the 0\% fully flash-backed setting, showing that it can be configured for different memory-constrained deployments rather than relying on a fixed amount of available DRAM.

\noindent\textbf{Output length.} Table~\ref{tab:eval-output-length} varies the maximum output length. Decode throughput remains stable across output lengths, with only a mild decrease for longer generations. This indicates that the benefit of \Scout{} is not limited to short responses; after prefill, each speculative cycle continues to amortize flash-backed target execution.

\section{Related Work}
\noindent\textbf{Memory-Constrained LLM Inference.}
Limited device memory has motivated model compression and hierarchy-aware inference. Quantization~\cite{gptq,smoothquant,awq,bitnet}, pruning, and sparsity~\cite{llm-pruner,dejavu,longexposure,jenga} reduce LLM memory or computation overhead, while KV-cache methods reduce runtime states through adaptive eviction, clustering, and storage-aware offloading~\cite{dynakv,mosaic,swarm}. Storage-aware systems such as FlexGen~\cite{flexgen} and LLM in a Flash~\cite{llm-flash} offload model tensors across memory and storage hierarchies. \Scout{} is complementary: instead of only fitting or streaming a large model, it reduces both the frequency and cost of flash-resident target invocations through speculative decoding.

\noindent\textbf{Speculative Decoding.} Speculative decoding accelerates autoregressive generation by using a lightweight draft model to propose tokens and a larger target model to verify them~\cite{speculative-decoding}. Prior work improves this process through tree verification, stronger draft models, and adaptive tree structures. SpecInfer verifies token trees in one target pass~\cite{specinfer}; Medusa generates candidates with multiple decoding heads~\cite{medusa}; EAGLE improves draft quality by predicting future features~\cite{eagle3}; and OPT-Tree optimizes draft-tree structures for higher expected acceptance~\cite{opttree}. These methods mainly assume that the target model is memory-resident and that verification can exploit server-class parallelism. In contrast, \Scout{} targets on-device LLM inference, where each verification step may incur flash I/O and mobile accelerators provide limited parallelism. \Scout{} therefore constructs token trees with a gain-cost objective that accounts for mobile constraints.

\noindent\textbf{On-Device LLM Inference.} On-device LLM systems and model designs exploit mobile-specific architectures, sparsity, storage, and heterogeneous compute~\cite{mobilellm}. Neuralink accelerates on-device inference using neuron co-activation~\cite{neuralink}, while PowerInfer-2 schedules LLM computation across smartphone CPU, NPU, and storage resources~\cite{powerinfer-2}. EdgeLLM applies speculative decoding to mobile LLM inference with branch navigation and verification pacing~\cite{edgellm}. \Scout{} differs by focusing on the DRAM-flash hierarchy in speculative decoding: drafting, verification, and execution are jointly optimized for this deployment setting.

\section{Conclusion}
We propose \Scout{}, a flash-backed speculative decoding system for smartphone LLM inference. By co-optimizing drafting, verification, and CPU-NPU execution, \Scout{} reduces mobile I/O and compute overheads under tight DRAM budgets. Our results show that \Scout{} enables efficient interactive inference for larger LLMs on smartphones.

\bibliographystyle{plain}
\bibliography{references}

\end{document}